\documentclass{article}
\usepackage{ijcai21}

\usepackage{times}

\usepackage{soul}
\usepackage{url}
\usepackage{hyperref}
\usepackage[all]{hypcap}
\hypersetup{
    colorlinks=true,
    linkcolor=blue,
    filecolor=magenta,      
    urlcolor=cyan,
}
\usepackage[utf8]{inputenc}
\usepackage[small]{caption}
\usepackage{graphicx}
\usepackage{amsmath}
\usepackage{amssymb}
\usepackage{tabularx} 
\usepackage{booktabs}
\usepackage{multirow}
\usepackage{multicol}
\usepackage{float}
\usepackage[ruled,vlined]{algorithm2e}
\usepackage{booktabs}

\title{A Benchmark and Empirical Analysis for Replay Strategies \\in Continual Learning}

\author{
Qihan Yang$^{1,2}$\and
Fan Feng$^1$\and
Rosa H.M. Chan$^{1}$\footnote{Contact Author}
\affiliations
$^1$City University of Hong Kong
\\
$^2$Imperial College London\\
\emails
yangqihan1999@gmail.com,
fan.feng@my.cityu.edu.hk,rosachan@cityu.edu.hk
}

\begin{document}

\maketitle

\begin{abstract}
With the capacity of continual learning, humans can continuously acquire knowledge throughout their lifespan. However, computational systems are not, in general, capable of learning tasks sequentially. This long-standing challenge for deep neural networks (DNNs) is called \textit{catastrophic forgetting}. Multiple solutions have been proposed to overcome this limitation. This paper makes an in-depth evaluation of the \textbf{memory replay methods}, exploring the efficiency, performance, and scalability of various sampling strategies when selecting replay data. All experiments are conducted on multiple datasets under various domains. Finally,  a practical solution for selecting replay methods for various data distributions is provided.
\end{abstract}

\section{Introduction}
\label{Introduction}

Unlike human beings who have continual learning ability, deep neural networks (DNNs), in general, suffer from a significant function decline or forgets the previously learned knowledge when encountering novel information and tasks. Extensive approaches have been proposed to overcome this limitation, including the regularization method (e.g., learning without Forgetting (LwF) \cite{LwF}, elastic weight consolidation (EWC) \cite{EWC}), dynamic architecture method (e.g., Progressive neural networks (PNN) \cite{pnn}), and the memory replay method (e.g., deep generative replay (DGR) \cite{DGR}). The regularization method regularizes the gradient descent path and tries to minimize the loss for new tasks while restricting the change of specific parameters to maintain the memory for the previous tasks. The dynamic architecture approach allocates new neural resources when the model is facing novel tasks. 
The memory replay method, proven to have the best performance \cite{icml2020}, rehearses the learned data from either a data buffer (i.e., experience replay) or a generative model (i.e., generative replay) and jointly trains the neural network with both old task and novel task data. 

Several benchmarks are proposed to evaluate the continual learning in computer vision tasks, including general computer vision~\cite{van2019three,lomonaco2017core50}, and robotic vision tasks~\cite{openloris}. However, these benchmarks only focus on specific datasets. Meanwhile, a wide range of continual learning methods is compared in these works, which are not equipped with the in-depth analysis of memory-based methods. In our work, we only focus on the analysis of replay-based methods, which are the most effective and practical methods, using different datasets. 
To provide this comprehensive benchmark, we evaluate  various memory-based replay strategies (including random, confidence, entropy, margin, K-means, core-set, maximally interfered retrieval (MIR), and Bayesian dis-agreement), replaying complex and simple data, and the difference between experience replay and generative replay. Using the standard continual learning datasets (i.e., MNIST \cite{MNIST}, CIFAR-10 \cite{CIFAR}, MiniImagenet \cite{imagenet}, and OpenLORIS-Object \cite{openloris}), we conduct $46$ experiments in total.

\section{Memory Replay}

Among all continual learning approaches, memorization-based approaches using experience replay  have generally proven more successful than their regularization-based alternatives \cite{icml2020}. There are, in general, two typical ways to get the replay data: \textit{experience replay} and \textit{generative replay}.

\subsection{Experience Replay}
\label{subsection:ExperienceReplay}
\begin{figure}[!hbt]
    \centering
    \includegraphics[width=8.5cm]{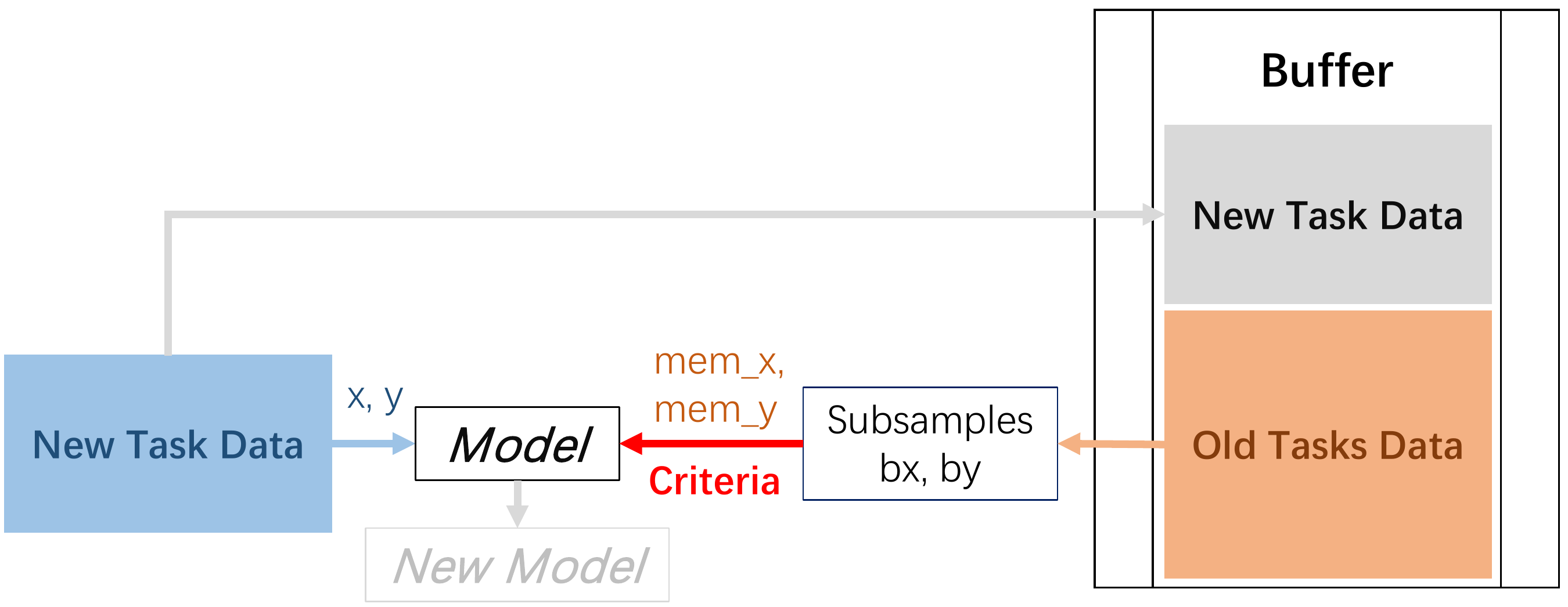}
    \caption{The workflow of experience replay.}
    \label{figure:experienceReplay}
\end{figure}

Early attempts of continual learning use the experience replay approach \cite{replay} (Figure~\ref{figure:experienceReplay}), which introduces a memory buffer that stores previous task data and regularly replay old samples interleaved with new task data. Specifically, when a new task comes with input data $x$ and labels $y$, the model will also load a subsample $(bx, by)$ from the buffer ($N_s$ represents the number of subsamples), and then a replay data $(\text{mem}\_x, \text{mem}\_y)$ is chosen from the subsamples based on \textit{certain criteria} ($N_r$ represents the number of replay data and $N_c$ represents the number of classes). The model performs training on both the current-task and the replay data. After training, the current task data will be stored in the buffer. As a result, the model can acquire the novel knowledge and maintain some previous memory simultaneously. 

\subsection{Generative Replay}
\begin{figure}[!hbt]
    \centering
    \includegraphics[width=8.5cm]{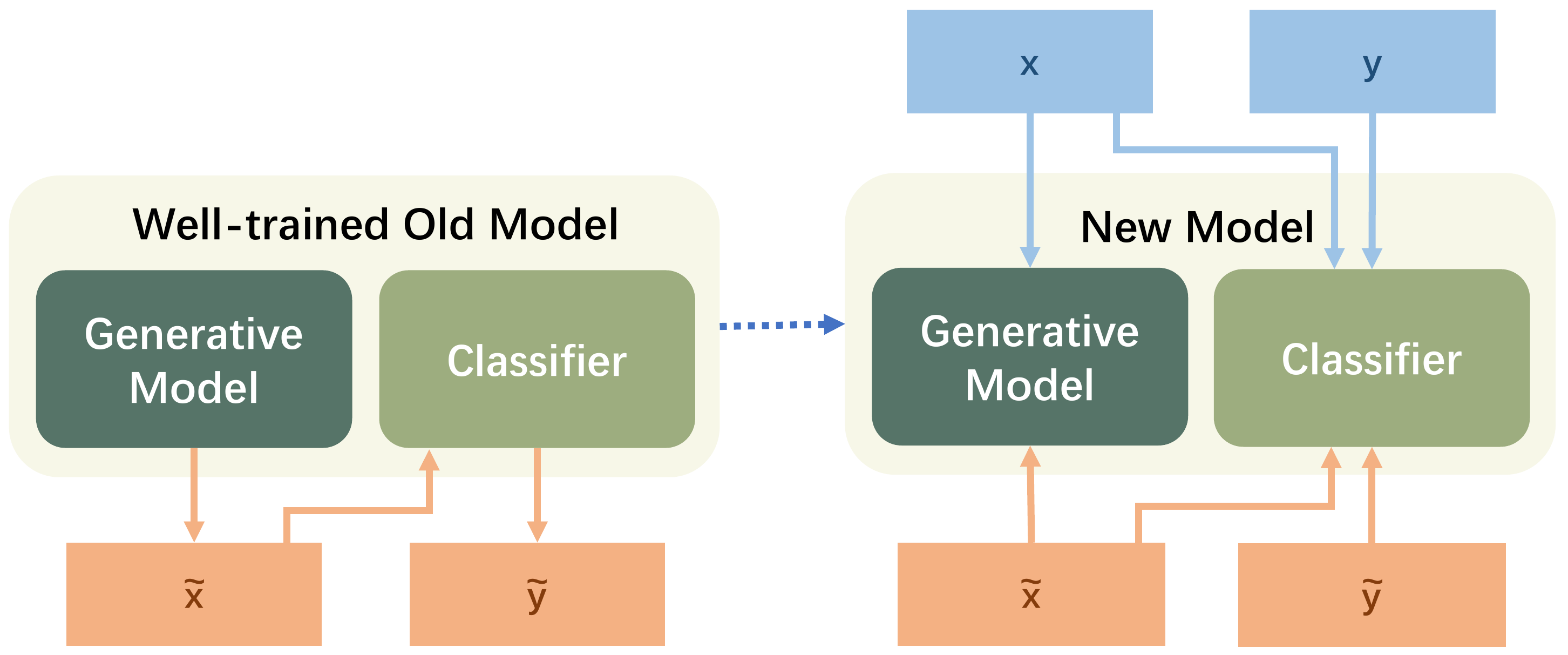}
    \caption{The workflow of generative replay.}
    \label{figure:generativeReplay}
\end{figure}

When the training data for old tasks are not available, only pseudo-inputs and pseudo-targets produced by a memory network can be fed into the model \cite{replay}. This rehearsal technique, named generative replay (Figure~\ref{figure:generativeReplay}), does not require extra memory storage but introduces a generative model parallel with the classifier, such as variational auto-encoder (VAE) \cite{VAE}, Generative adversarial network (GAN) \cite{GAN}, and Wasserstein GAN (WGAN) \cite{WGAN}. At any training stage, the generative model can produce pseudo data $\tilde{x}$, which imitates the old data that have been trained; the classifier can correctly classify the tasks that have been traversed and give pseudo prediction $\tilde{y}$ based on $\tilde{x}$. When the new task comes with input data $x$ and labels $y$, both the generative model and the classifier are jointly trained with ($\tilde{x}$, $\tilde{y}$) and ($x$, $y$). After training, the generative model will be capable of producing pseudo data that mimic both the previous tasks and the current task data; the classifier will be able to classify both the old tasks and novel task correctly.

\section{Evaluation}

This paper evaluates $8$ replay strategies, which can be classified as random selection, feature embedding, model prediction, and maximal interference. 
\begin{itemize}
    \item \textit{Model Prediction}: entropy, confidence, margin, Bayesian dis-agreement, and MIR;
    \item \textit{Feature Embedding}: K-means and core-set.
    \item \textit{Random sampling}.
\end{itemize}

With the random replay (randomly choosing $N_r$ replay data from $N_s$ subsamples) as a benchmark, we explored the open problems with multiple datasets under various domains.
\begin{enumerate}
    \item the efficiency, performance, and scalability of different sampling strategies when selecting replay data;
    \item the relationship between the efficiency and the difficulty of replay data;
    \item the performance gap between experience replay and generative replay methods;
    \item the relationship between the performances and continual learning orders. 
\end{enumerate}
Based on the empirical findings and analysis, we also provide a practical solution for selecting replay methods for datasets under different domains.

\subsection{Feature Embedding}
We use ResNet-18 \cite{ResNet} in the experience replay methods. For simplicity, the output features of layer$1$, layer$2$, and layer$3$ are modified to $20$, $40$, and $80$. The current classifier takes the subsamples $\boldsymbol{bx} = [bx^1, \dots, bx^{N_s}]$ as input, and the output of layer$3$ will be extracted as feature maps. Notice the dimension of these maps is $(N_s, 80, \star, \star)$. Then, a 2D adaptive average pooling layer will extract the feature vector with dimension $(N_s, 80, 1, 1)$ from the feature maps, followed by a flatten layer stretches the feature vectors to the embeddings $\boldsymbol{e} = [e^1, \dots, e^{N_s}]$ with each ${e}^i$ for $i \in \{1, \dots, N_s\}$  has a dimension of $(N_s, 80)$. After getting the embeddings for subsamples, the replay data can be selected from it with different algorithms (embeddings' \textbf{K-means} and \textbf{core-set}). 

\begin{itemize}
    \item \textbf{K-means}: K-means is one of the most well-known clustering method \cite{Kmeans}. With the $N_s$ embeddings, it performs a $N_r$ center clustering. Then, from each cluster, the embedding closest to the centre is the most representative point, which will be chosen to be the replay data $(\text{mem}\_x, \text{mem}\_y)$.
    
    \item \textbf{core-set}: Core-set is a small set of points that approximate a large group of the point's shape and distribution \cite{coreset1,coreset2,campbell2019sparse}. Here, we use \textit{Max-Min Distance} to select the core-set. Detailed procedures are shown in algorithm \ref{algorithm:minmax}.
    
    \begin{algorithm}
        \SetAlgoLined
        \KwResult{Get $N_r$ representative replay data from $N_s$ subsamples}
        \textbf{Input}: Randomly select an embedding $e^i$ as the initial point, $i \in \{1, \dots, N_s\}$; mark the initial point as \textit{"selected"} point; mark all the other embeddings as \textit{"unselected"} points.
        
        \Repeat{$N_r$ embeddings have been marked as \textit{"selected"}}{
        \begin{itemize}
            \item For all the \textit{unselected} points, cluster them to the \textit{selected} points based on the minimal distance\;
            \item Select the embedding with the maximum minimal distance as the subsequent replay data\;
            \item Mark the new embedding as the \textit{selected} point.
        \end{itemize}
         }
         \caption{Max-Min Distance Algorithm}
         \label{algorithm:minmax}
    \end{algorithm}
\end{itemize}

\subsection{Model Prediction}
Besides strategies based on feature embedding, there are four popular strategies based on the model prediction. The current model takes the subsamples as input, performs an evaluation, and gets the predictions with size $(N_s, N_c)$. For any subsample ${bx}^i$ for $i \in \{1, \dots, N_s\}$, the prediction is given $\hat{by}^i = [\hat{by}_{1}^{i}, \dots, \hat{by}_{N_c}^{i}]$. With this prediction, the replay data $(mem\_x, mem\_y)$ can be selected based on different calculation methods (\textbf{confidence}, \textbf{entropy}, \textbf{margin}, and \textbf{Bayesian dis-agreement}).

\begin{itemize}
    \item \textbf{Confidence (C)}: For any subsample ${bx}^i$ for $i \in \{1, \dots, {N_s}\}$, the \textit{Confidence} $C^i$ is given by its largest class prediction $\hat{by}_{max}^{i}$ among all the $\hat{by}_{c}^{i}$ for $c \in \{1, \dots, {N_c}\}$ (equation \ref{equation:Confi}). A large confidence stands for simple replay data and a small confidence stands for difficult replay data.
    \begin{align}
        C^{i}\left(\hat{by}^i \mid {bx}^i\right) = \hat{by}_{max}^{i} = \max \left(\hat{by}_{1}^{i}, \dots, \hat{by}_{N_c}^{i}\right)
        \label{equation:Confi}        
    \end{align}
    \begin{itemize}
        \item Simple replay: top $N_r$ largest $C^i$, $i \in \{1, \dots, N_s\}$.
        \item Difficult replay: top $N_r$ smallest $C^i$, $i \in \{1, \dots, N_s\}$.
    \end{itemize}
    
    \item \textbf{Entropy (H)}: Entropy denotes the \textit{randomness} or the \textit{uncertainty} of the data. From the model prediction $\hat{by}^i = [\hat{by}_{1}^{i}, \dots, \hat{by}_{N_c}^{i}]$, the \textit{Entropy} $H^i$ can be calculated with equation \ref{equation:Entropy} for any subsample ${bx}^i$ for $ i \in \{1, \dots, {N_s}\}$. Similarly, a small entropy stands for high certainty (simple) replay data and a large entropy stands for low certainty (difficult) replay data \cite{Confi/Entropy}.
    \begin{equation}
        H^i\left(\hat{by}^i \mid {bx}^i\right) = -\sum_{c=1}^{N_c} \left(\hat{by}_{c}^{i} \times \log \hat{by}_{c}^{i}\right)
    \label{equation:Entropy}
    \end{equation}
    \begin{itemize}
        \item Simple replay: top $N_r$ smallest $H^i$, $i \in \{1, \dots, N_s\}$.
        \item Difficult replay: top $N_r$ largest $H^i$, $i \in \{1, \dots, N_s\}$.
    \end{itemize}
    
    \item \textbf{Margin (M)}: For any subsample ${bx}^i$ for $ i \in \{1, \dots, {N_s}\}$, with its prediction $\hat{by}^i = [\hat{by}_{1}^{i}, \dots, \hat{by}_{N_c}^{i}]$, the \textit{Margin} $M^i$ is given by difference between the largest class prediction $\hat{by}_{max}^{i}$ and the smallest class prediction $\hat{by}_{min}^{i}$ (equation \ref{equation:Margin}). A large margin stands for simple replay data, and a small margin stands for difficult replay data \cite{margin}.
    \begin{align}
        \label{equation:Margin}
        M^{i}\left(\hat{by}^i \mid {bx}^i\right) = \hat{by}_{max}^{i} - \hat{by}_{min}^{i}\\
        \hat{by}_{max}^{i} = \max \left(\hat{by}_{1}^{i}, \dots, \hat{by}_{N_c}^{i}\right) \notag\\
        \hat{by}_{min}^{i} = \min \left(\hat{by}_{1}^{i}, \dots, \hat{by}_{N_c}^{i}\right) \notag
    \end{align}
    \begin{itemize}
        \item Simple replay: top $N_r$ largest $M^i$, $i \in \{1, \dots, N_s\}$.
        \item Difficult replay: top $N_r$ smallest $M^i$, $i \in \{1, \dots, N_s\}$.
    \end{itemize}
    
    \item \textbf{Bayesian Dis-agreement (B)}: Bayesian dis-agreement strategy is inherited from the \textit{(Bayesian Active Learning by Dis-agreement} (BALD) \cite{Bayesian}. This method takes the assumption that there are some model parameters control the dependence between inputs and outputs. As a result, an acquisition function is defined to estimate the mutual information between the model parameters and the model predictions. In other words, it captures how strongly the model parameters and the model predictions for a given data point are coupled \cite{bald}. 
    
    Given a model $\mathcal{M}$ with parameter $\boldsymbol{\omega}$, a training set $\mathcal{D}_\text{train}$, the Bayesian dis-agreement $B^i$ for a subsample ${bx}^i$ for $ i \in \{1, \dots, {N_s}\}$ is calculated with equation \ref{equation:bayesian}.
    
    \begin{align}
    {B}^i\left(\hat{by}^i, \boldsymbol{\omega} \mid {bx}^i, \mathcal{D}_{\text {train }}\right) = {H}\left(\hat{by}^i \mid {bx}^i, \mathcal{D}_{\text {train}}\right) \notag \\ -\mathbb{E}_{p\left(\boldsymbol{\omega} \mid \mathcal{D}_{\text {train }}\right)}[{H}(\hat{by}^i \mid {bx}^i, \boldsymbol{\omega})] 
    \label{equation:bayesian}
    \end{align}
    
    The first term ${H}\left(\hat{by}^i \mid {bx}^i, \mathcal{D}_{\text {train}}\right)$ is the \textit{Entropy} of the model prediction. For given dataset, the first term is small if the model prediction is certain. The second term $\mathbb{E}_{p\left(\boldsymbol{\omega} \mid \mathcal{D}_{\text {train }}\right)}[{H}(\hat{by}^i \mid {bx}^i, \boldsymbol{\omega})$ is an expectation of the entropy of the model prediction over the posterior of the model parameters. For each draw of model parameters from the posterior, the second term is small if the model prediction is certain \cite{batchBald}. 
    
    In general, the input will have a large $B$ if its model prediction is prone to vary with changing model parameters and vice versa. A small Bayesian dis-agreement represents stable (simple) replay data, and a large bayesian dis-agreement represents unstable (difficult) replay data.
    \begin{itemize}
        \item Simple replay: top $N_r$ smallest $B^i$, $i \in \{1, \dots, N_s\}$.
        \item Difficult replay: top $N_r$ largest $B^i$, $i \in \{1, \dots, N_s\}$.
    \end{itemize}
    
For any subsample, it will get $k$ predictions based on distinct parameters with the help of a drop out layer; then the first term of the acquisition function can be calculated as the entropy of the mean of the $k$ predictions, and the second term can be calculated as the mean of the $k$ entropy.
\end{itemize}

\begin{figure}[!hbt]
    \centering
    \includegraphics[width=8.2cm]{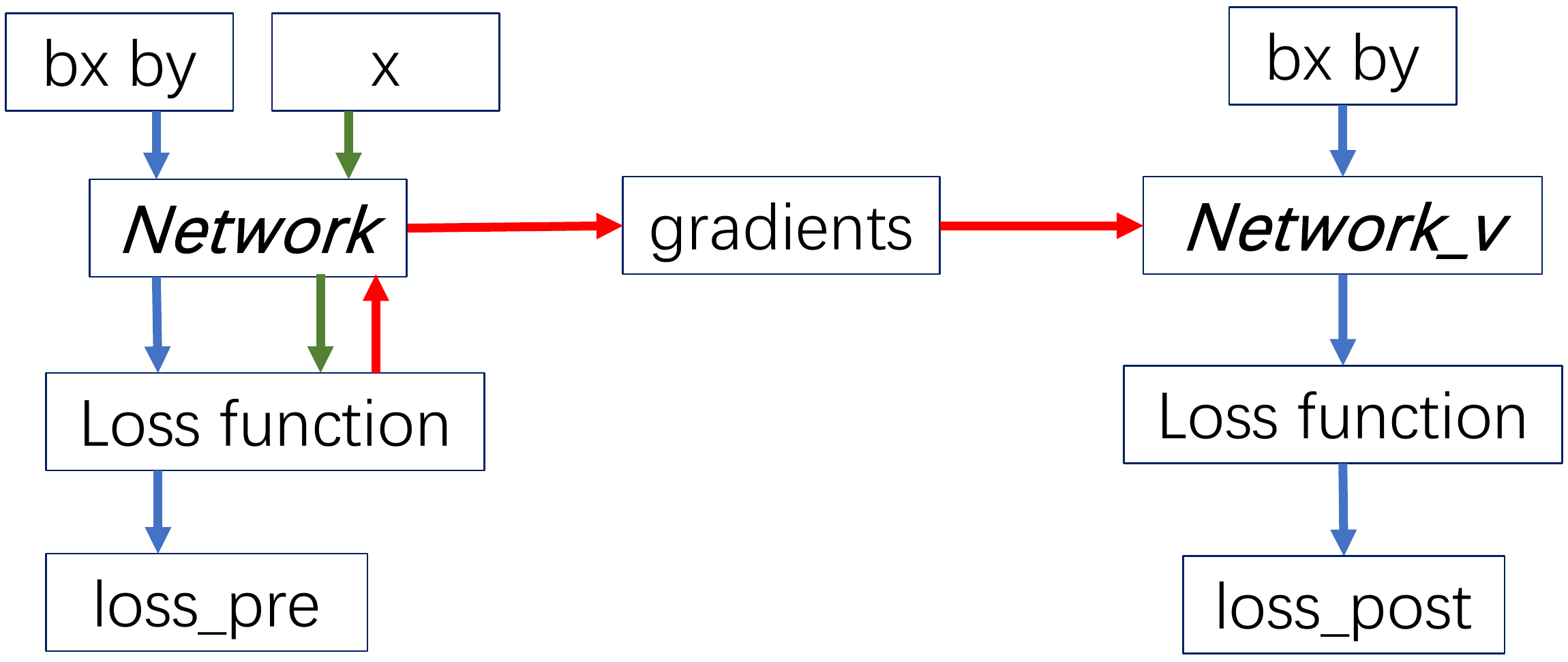}
    \caption[Maximal Interfered Retrieval strategy demonstration]{The workflow of MIR: the green arrows denote the forward propagation of new data x; the red arrows denote the back propagation of new data; the blue arrows denote the prediction procedure of subsamples.}
    \label{figure:mir}
\end{figure}

The strategies that rely on feature embeddings and model predictions manage to study the properties of sampled data points and select some representative points to form the replay data. However, they do not take the influence of the new coming data into account. MIR (Figure~\ref{figure:mir}), on the other hand, is a replay strategy based on the novel task data's interference.

\subsection{Maximal Interfered Retrieval (MIR)}
\label{subsection:MIR}

The crucial idea of MIR is to filter out the samples which will be maximally interfered by the new task data \cite{mir}. The interference is measured by the difference between the loss value before and after training the novel data. To get the loss after training the new task data, MIR performs a \textit{virtual update} according to the gradients generated during training the novel data.

The loss function of the classifier is represented by equation \ref{equation:lossFunc}, and the virtual update can be seen as equation \ref{equation:virtualUpdate}. 
\begin{equation}
    \mathcal{L}\left(\mathcal{M}_{\omega}\left(\boldsymbol{x}\right), \boldsymbol{y}\right)
    \label{equation:lossFunc}
\end{equation}

\begin{equation}
    \omega^{v} = \omega - \eta\nabla\mathcal{L}\left(\mathcal{M}_{\omega}\left(\boldsymbol{x}\right), \boldsymbol{y}\right)
    \label{equation:virtualUpdate}
\end{equation}

For any subsample ${bx}^i$ for $ i \in \{1, \dots, {N_s}\}$, with its prediction $\hat{by}^i$, current model $\mathcal{M}_{\omega}$ and virtual updated model $\mathcal{M}_{\omega^{v}}$, the MIR score $S^i$ is calculated with equation \ref{equation:mir}.
\begin{align}
    S^i &=\mathcal{L}^{v} - \mathcal{L} \notag\\
    &= \mathcal{L}\left(\mathcal{M}_{\omega^{v}}\left({bx}^i\right), {by}^i\right) - \mathcal{L}\left(\mathcal{M}_{\omega}\left({bx}^i\right), {by}^i\right)
    \label{equation:mir}
\end{align}

A lower score means that the subsample is less affected by the new data; a higher score represents that the subsample is more affected by the new data, and those high score data will be selected as replay data. 

The detailed algorithm is shown in algorithm \ref{algorithm:mir}.

\begin{algorithm}
\SetAlgoLined
\KwResult{Get $N_r$ most interfered replay data from $N_s$ subsamples.}
 \textbf{Input}: new task data $(\boldsymbol{x}, \boldsymbol{y})$, subsamples $(\boldsymbol{bx}, \boldsymbol{by})$\;
 \textbf{Given}: Model $\mathcal{M}$ parameters $\omega$\;
 \begin{itemize}
    \item Train the model $\mathcal{M}_{\omega}$ with new data $(\boldsymbol{x}, \boldsymbol{y})$ and record the gradients $\nabla\mathcal{L}$;
    \item Copy the current model to a new model, perform virtual update based on the gradients, and get the virtual model $\mathcal{M}_{\omega^{v}}$;
    \item Evaluate the subsamples $(\boldsymbol{bx}, \boldsymbol{by})$ with both the current model and the virtual model, get two losses $\mathcal{L}$, $ \mathcal{L}^{v}$;
    \item Calculate the MIR score $\boldsymbol{S} = \mathcal{L}^{v} - \mathcal{L}$;
    \item Select replay data: $\left(\boldsymbol{mem\_x}, \boldsymbol{mem\_y}\right)$ is the $\left(\boldsymbol{bx}, \boldsymbol{by}\right)$ with the top $N_r$ largest $S^i$, for $i \in \{1, \dots, N_s\}$.
 \end{itemize}
 \caption{Maximal Interfered Retrieval algorithm}
 \label{algorithm:mir}
\end{algorithm}

\subsection{Datasets}
The datasets used in this project are \textbf{MNIST}, \textbf{CIFAR-10}, \textbf{MiniImagenet}, and \textbf{OpenLORIS-Object}. Each of them is separated into several tasks.
\begin{itemize}
    \item \textbf{MNIST}:  separated into ${5}$ disjointed tasks. For each task, $1,000$ samples are used for training.
    
    \item \textbf{CIFAR-10}: separated into ${5}$ disjointed tasks. Each task consists of ${2}$ non-overlapped classes with $9,750$ training data, $250$ validation data, and $2,000$ test data.
    
    \item \textbf{MiniImagenet}: divided into ${20}$ disjointed tasks with ${5}$ categories each. Each task consists of $2400$ training images.
    
    \item \textbf{OpenLORIS-Object}: separated into four tasks based on the four factors, and each task has all  $121$ classes.
\end{itemize}

\section{Results and Discussion}

\subsection{Evaluation Criteria}
Three evaluation criteria are used to denote the model performance. Assume a given dataset $\mathcal{D}$ has $T$ tasks $\mathcal{T}_1, \dots, \mathcal{T}_T$. After training the last task $\mathcal{T}^T$, the \textit{Accuracy} and \textit{Forgetting rate} is given by equation \ref{equation:accuracy} and equation \ref{equation:froget}. The $Acc_{\mathcal{T}_i}$ represents the accuracy for task $\mathcal{T}_i$ after training the last task, and the $Best_{\mathcal{T}_i}$ represents the best accuracy for task $\mathcal{T}_i$ during the whole training procedure. Usually the $Best_{\mathcal{T}_i}$ is achieved right after the $\mathcal{T}_i$ just be trained.

\begin{equation}
    \text{Accuracy} = \frac{1}{T} \sum_{i=1}^{T} \text{Acc}_{\mathcal{T}_i}
    \label{equation:accuracy}
\end{equation}
\begin{equation}
    \text{Forgetting Rate} = \frac{1}{T} \sum_{i=1}^{T} \left( \text{Best}_{\mathcal{T}_i} - \text{Acc}_{\mathcal{T}_i} \right)
    \label{equation:froget}
\end{equation}

The third criteria \textit{Average Time} is the average running time for a single run. Given total running time $T_r$ and number of runs $n_r$, the average time ($s$) is calculated by equation \ref{equation:avetime}
\begin{equation}
    \text{Average Time} = \frac{T_r}{n_r} 
    \label{equation:avetime}
\end{equation}

\subsection{Replay Strategies Comparison}
When comparing different replay strategies, all other factors are set to default: experience replay with simple data. Eight replay strategies are trained with the same settings under same conditions, and compared parallelly. The detailed results are shown in appendix table \ref{table:Sum1}. The \textbf{highest accuracy}, the \textbf{lowest forgetting rate}, and the \textbf{minimal running time} (besides random replay) are embolden. The recordings of the \textit{maximal running time} are italic. Table \ref{table:summary} and \ref{table:strategies} summarize the results.

\begin{table}[ht]
\centering
\caption{Highest accuracy, lowest forgetting rate, least average time, and largest average time achieved on each dataset.}
\label{table:summary}
\resizebox{8.5cm}{!}{
\begin{tabular}{@{}lcccc@{}}
\toprule
\textbf{} & \textbf{Accuracy} & \textbf{\begin{tabular}[c]{@{}c@{}}Forgetting\\ Rate\end{tabular}} & \textbf{\begin{tabular}[c]{@{}c@{}}Least Average\\ Time (s)\end{tabular}} & \textbf{\begin{tabular}[c]{@{}c@{}}Largest Average\\ Time (s)\end{tabular}} \\ \midrule\midrule
\textbf{MNIST} & $0.876$ & $0.096$ & $3.700$ & $22.400$ \\
\textbf{CIFAR-10} & $0.409$ & $0.221$ & $89.200$ & $282.933$ \\
\textbf{MiniImagenet} & $0.158$ & $0.280$ & $1284.400$ & $1745.800$ \\
\textbf{OpenLORIS} & $0.966$ & $0.014$ & $5777.000$ & $12270.400$ \\ \bottomrule
\end{tabular}}
\end{table}

\begin{table}[ht]
\caption[Strategies' results summary]{Best performances of different replay strategies on the CIFAR-10, MiniImagenet and OpenLORIS-Object.}
\label{table:strategies}
\centering
\resizebox{8.5cm}{!}{
\begin{tabular}{@{}lcccc@{}}
\toprule
\textbf{} &
  \textbf{Accuracy} &
  \textbf{\begin{tabular}[c]{@{}c@{}}Forgetting\\ Rate\end{tabular}} &
  \textbf{\begin{tabular}[c]{@{}c@{}}Least Average\\ Time (s)\end{tabular}} &
  \textbf{\begin{tabular}[c]{@{}c@{}}Largest Average\\ Time (s)\end{tabular}} \\ \midrule \midrule
  \textbf{CIFAR-10} &
  MIR &
  \begin{tabular}[c]{@{}c@{}}Bayesian\end{tabular} &
  Entropy &
  K-means \\
  \textbf{MiniImagenet} &
  \begin{tabular}[c]{@{}c@{}}Bayesian\end{tabular} &
  \begin{tabular}[c]{@{}c@{}}Bayesian\end{tabular} &
  \begin{tabular}[c]{@{}c@{}}C. \& M.\end{tabular} &
  K-means \\
  \textbf{OpenLORIS} &
  \begin{tabular}[c]{@{}c@{}}H. \& C.\end{tabular} &
  Entropy &
  Entropy &
  \begin{tabular}[c]{@{}c@{}}Bayesian\end{tabular} \\ \bottomrule
\end{tabular}}
\end{table}

For MNIST, almost all the replay strategies achieve a similar result, and the performances are, by and large, reasonably good (accuracy $\textbf{0.870}$, forgetting rate $\textbf{0.100}$). The reason behind this might be that MNIST is too simple, and no matter choose what kind of replay data, even using random selected data, makes no significant difference. However, for a slightly diverse dataset like CIFAR-10, the general accuracy is lower, and the forgetting rate is higher (accuracy between $\textbf{0.339}$ to $\textbf{0.409}$ and a forgetting rate between $\textbf{0.221}$ to $\textbf{0.400}$). For more complex dataset like MiniImagenet, the accuracy drop exponentially, and the forgetting problem is more severe (accuracy between $\textbf{0.107}$ to $\textbf{0.158}$ and a forgetting rate between $\textbf{0.280}$ to $\textbf{0.402}$). OpenLORIS-Object also has complex data distribution, but it achieves an accuracy between $\textbf{0.960}$ to $\textbf{0.966}$ and a forgetting rate between $\textbf{0.014}$ to $\textbf{0.022}$. Nevertheless, this outstanding performance is not comparable to the other datasets due to the task separation difference.

In the continual learning field, there are different \textit{Incremental Learning Scenarios} as follow:
\begin{itemize}
    \item \textbf{Instance incremental}: The class number is fixed, while in each learning stage, more novel instances are involved. 
    \item \textbf{Class-incremental}: New classes are introduced in each learning stage.
    \item \textbf{Domain-incremental}: The class number is also fixed; however, the model will be used in the context of a different but related input distribution.
\end{itemize}

The tasks for MNIST, CIFAR-10, and MiniImagenet are in a class-incremental manner. Each time the classifier will encounter new categories that it has not seen before. However, for OpenLORIS-Object, with all classes being presented in the first task, the following tasks are trying to perform similar classification to the same categories under different environmental factors. This follows the domain-incremental manner. The difficulty for training a class-incremental model is much greater than training a domain-incremental model. Thus, the forgetting rate for OpenLORIS-Object is generally lower than that in the other three datasets.

For CIFAR-10 and MiniImagenet and OpenLORIS-Object, different replay strategies show their uniqueness.
\begin{itemize}
    \item \textbf{Bayesian Dis-agreement}: Bayesian dis-agreement has an outstanding performance on the CIFAR-10 and MiniImagenet; however, its forgetting rate on the OpenLORIS-Object increases dramatically compared with other strategies. The class number could be a problem. CIFAR-10 only has $\textbf{2}$ classes per task ($5$ tasks and $10$ categories in total), and MiniImagenet has $\textbf{5}$ classes per task ($20$ tasks and $100$ categories in total). However, OpenLORIS-Object includes all the $\textbf{121}$ classes for each task. This outcome implies that the Bayesian dis-agreement strategy has good performance with a small class number, but it has \textbf{low scalability}. 
    
    Intuitively, when the class number gets larger, the mutual information between the model parameters and output prediction will be more complex and ambiguous; thus, the Bayesian dis-agreement strategy will fail to select the representative replay data based on that mutual information.
    
    \item \textbf{Entropy, confidence, and margin}: These three methods have neutral performances for any dataset. Since they select the replay data only based on the model prediction,  the results are more stable and \textbf{plastic}: They neither have a satisfying result as Bayesian dis-agreement on CIFAR-10 and MiniImagenet nor suffer performance decay when the class number gets more extensive as in OpenLORIS-Object. Moreover, the training time for these three methods, in general, are less than Bayesian dis-agreement, K-means, and MIR. As a result, for complex datasets with large data scale and class number, these three methods might be the most efficient choices. 
    
    \item \textbf{MIR}: Theoretically, MIR should be the best method since it tries to replay the data that interfered the most with the new coming data. Practically, it does not show its overwhelming advantage. 

    On the one hand, the subsample size $N_s$ may not be sufficient for MIR to traverse enough learned data and determine the most interfered samples. On the other hand, the prior assumption for MIR - replay the data with the largest loss difference between before and after training the new data will reinforce the old memory - is reasonably correct, but may not be as efficient as studying the model output directly as entropy, confidence, and margin strategies. It also takes time to evaluate the interference by making the prediction twice with both the current model and virtual updated model. Consequently, MIR strategy producing limited improvement, consumes more time than the model prediction-based strategies.
    
    \item \textbf{K-means and core-Set}: K-means method is very time consuming, and it has a slight improvement when the data is simple and the class number is low. Similarly, a small class number and the limited number of subsamples may not be sufficient for K-means to provide meaningful clusters and discover valuable replay data. core-Set strategy is also optimal considering its running time and the optimization effect it has.
\end{itemize}

\subsection{Replay Data Difficulty Comparison}
Entropy, confidence, margin, and Bayesian dis-agreement can be used to compare the difference between replay simple data and replay difficult data. Detailed results are shown in appendix table \ref{table:Sum2}. The \textbf{better performance} in accuracy (higher), forgetting rate (lower) and average time (shorter) are embolden. For MNIST, replay simple data and replay difficult data does not have clear distinctions. However, for CIFAR-10 and MiniImagenet, replaying difficult data generally lead to higher accuracy and lower forgetting rate. The running time for the two cases are comparable.

\subsection{Experience \& Generative Replay Comparison}
\begin{figure}[!hbt]
    \centering
    \includegraphics[height=1.6cm]{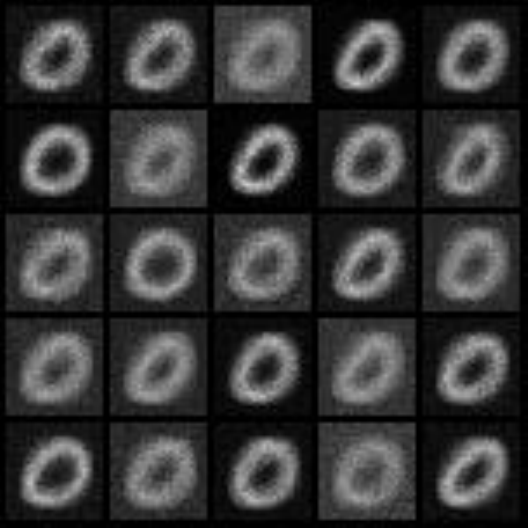}
    \includegraphics[height=1.6cm]{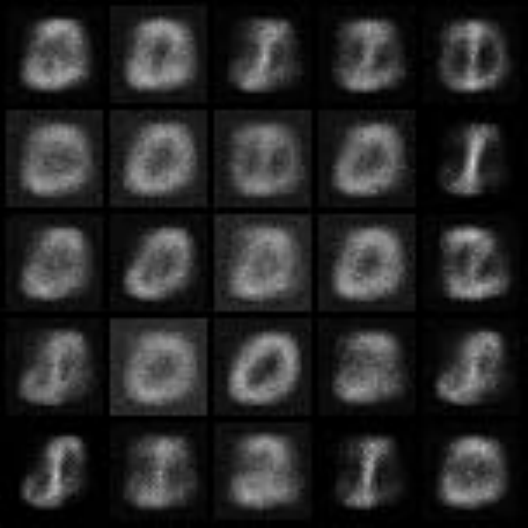}
    \includegraphics[height=1.6cm]{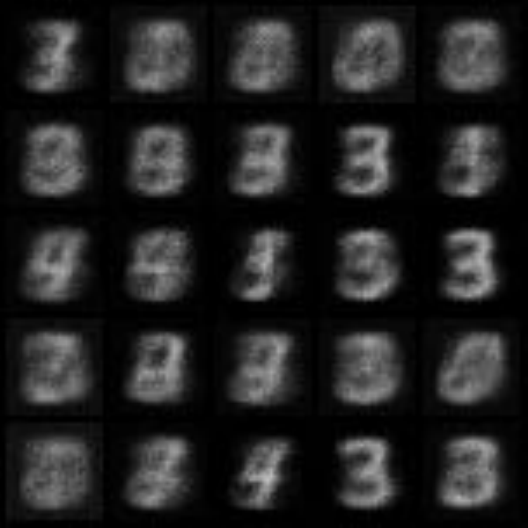}
    \includegraphics[height=1.6cm]{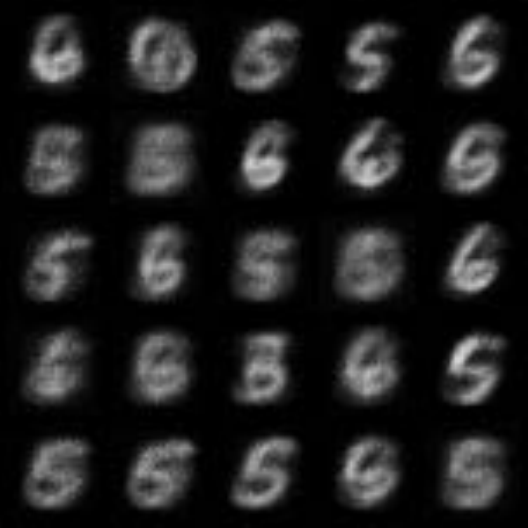}
    \includegraphics[height=1.6cm]{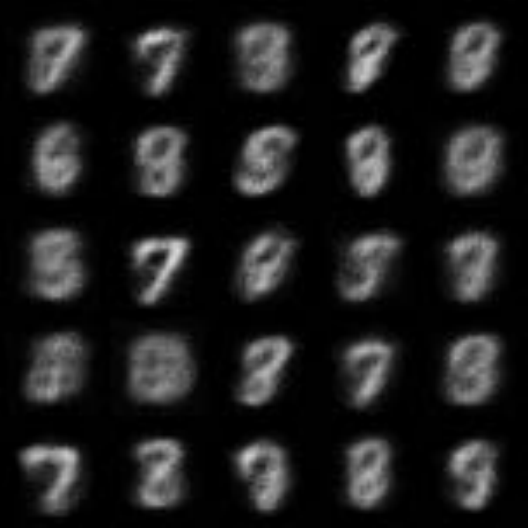}
    \caption{The replay data samples for each task.}
    \label{figure:replaydata}
\end{figure}

\begin{figure}[!hbt]
    \centering
    \includegraphics[width=4.2cm]{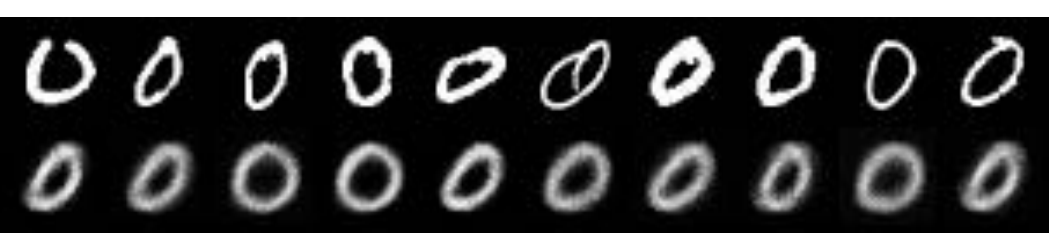}
    \includegraphics[width=4.2cm]{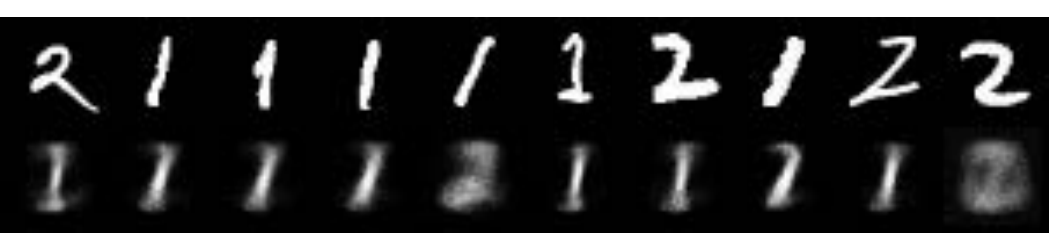}
    \includegraphics[width=4.2cm]{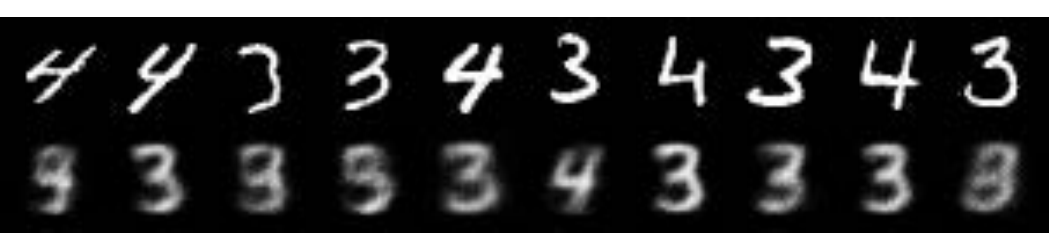}
    \includegraphics[width=4.2cm]{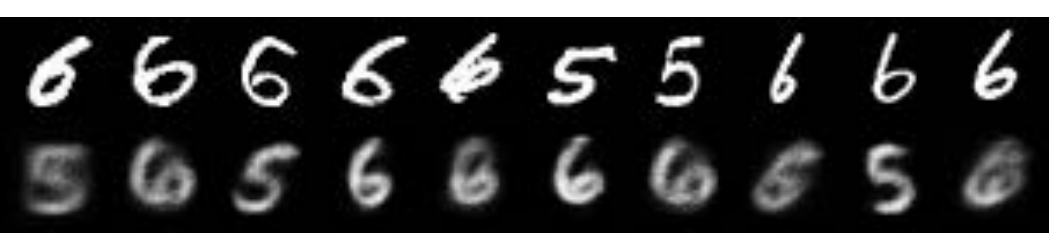}
    \includegraphics[width=4.2cm]{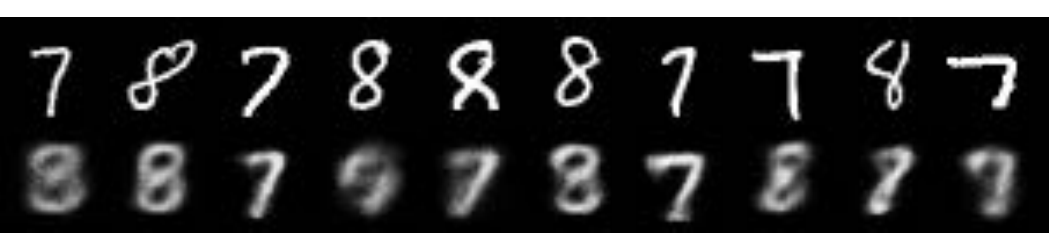}
    \caption{The reconstruction image samples for the five tasks.}
    \label{figure:reconstruct}
\end{figure}

MNIST is used to compare the efficiency of experience replay and generative replay. The generative replay uses VAE model \cite{VAE}. The results for experience replay and generative replay comparison are shown in the table \ref{table:MNIST_3}. The replay data samples during training each task and the reconstruction images for $5$ tasks are shown in figure \ref{figure:replaydata} and \ref{figure:reconstruct}.

\begin{table}[h]
\centering
\caption{MNIST ER and GR comparison results.}
\label{table:MNIST_3}
\resizebox{8.2cm}{!}{
\begin{tabular}{@{}cccc@{}}
\toprule
\multicolumn{2}{c}{\textbf{MNIST}} & \textbf{Experience Replay} & \textbf{Generative Replay} \\ \midrule \midrule
\textbf{} & Accuracy & $\textbf{0.870}$ +/- $\textbf{0.012}$ & $0.563$ +/- $0.005$ \\
\textbf{\begin{tabular}[c]{@{}c@{}}Evaluation\\ Criteria\end{tabular}} & \begin{tabular}[c]{@{}c@{}}Forgetting\\ rate\end{tabular} & $\textbf{0.105}$ +/- $\textbf{0.015}$ & $0.486$ +/- $0.007$ \\
\textbf{} & \begin{tabular}[c]{@{}c@{}}Average time\\ (s)\end{tabular} & $\textbf{3.050}$ & $6.754$ \\ \bottomrule
\end{tabular}}
\end{table}

According to the result, the experience replay has higher accuracy, lower forgetting rate, and less time than the generative replay. On the one hand, the generated data do not faithfully imitate the original image due to the replay model's imperfections. On the other hand, training the generative model introduces extra time consumption. Following the conclusion in \cite{generativeProblem}, the generative replay has worse performance on complex dataset such as CIFAR-10, and continual learning with generative replay remains a challenge.
\subsection{Training Sequence Comparison}
Since the OpenLORIS-Object dataset splits the data difficulty into three levels, the difference between training the hard data (weak illumination, high occlusion rate, small object size or longer distance from the camera, complex environment) first, and training the easy data (strong illumination, low occlusion rate, big object size or short distance from the camera, simple environment) first could be explored and evaluated. The result is shown in table \ref{table:OpenLORIS_3}. According to the result, learn the data with small environmental interfered (low difficulty level) and simple data distribution first, then moving to the challenging training samples (high difficulty level) helps the model get higher final accuracy and lower forgetting rate.
\begin{table}[h]
\centering
\caption{OpenLORIS-Object training sequence comparison results.}
\label{table:OpenLORIS_3}
\resizebox{8cm}{!}{
\begin{tabular}{@{}cccc@{}}
\toprule
\multicolumn{2}{c}{\textbf{OpenLORIS}} & \textbf{Easy First} & \textbf{Hard First} \\ \midrule
\textbf{} & Accuracy & $\textbf{0.096}$ +/- $\textbf{0.000}$ & $0.095$ +/- $0.016$ \\
\textbf{\begin{tabular}[c]{@{}c@{}}Evaluation\\ Criteria\end{tabular}} & \begin{tabular}[c]{@{}c@{}}Forgetting\\ rate\end{tabular} & $\textbf{0.022}$ +/- $\textbf{0.007}$ & $0.034$ +/- $0.020$ \\
\textbf{} & \begin{tabular}[c]{@{}c@{}}Average time\\ (s)\end{tabular} & $5521.800$ & $\textbf{5294.400}$ \\ \bottomrule
\end{tabular}}
\end{table}

\subsection{Analysis}

Based on the experimental results\footnote{All the methods are implemented using PyTorch~\cite{pytorch} toolbox with an Intel Core i9 CPU and $4$ Nvidia RTX 2080 Ti GPUs.}, we draw the concluding remarks: 
\begin{itemize}
    \item With the same setting, experience replay can achieve better accuracy and a lower forgetting rate compared with generative replay, but it needs an extra working memory to store all previous data. This may lead to memory problem when the task number gets larger or the scale of the dataset is large.
    \item Under experience replay, for a simple dataset like MNIST, different replay strategies make no significant difference, and they all achieve reasonable good accuracy and a low forgetting rate. For slightly complex datasets like CIFAR-10 or even more challenging dataset like MiniImagenet, Bayesian dis-agreement leads to a good forgetting rate when the categories per task are small. However, Bayesian dis-agreement has low scalability, and it is relatively time-consuming.
    \item MIR improves the performance compared to random replay, but it also takes more time to train than model prediction-based methods. Moreover, it is not guaranteed to get the lowest forgetting rate or the best accuracy. Embedding based methods like K-means and core-set are not ideal under the same training condition compared with other methods. Thus, for a complex dataset with a large class number for any single take, simple replay strategies like entropy, confidence, and margin are the most effective approaches.
    \item Replay difficult data is, in general, more optimal for the continual learning model to achieve a low forgetting rate. For OpenLORIS-Object, train the model with objects under simple environmental factors first slightly improves the model performance. 
\end{itemize}
\section{Conclusion}
This paper provides a benchmark and empirical analysis on the  replay-based methods, which are theoretically proven to be the best-performing methods in continual learning~\cite{icml2020}. The extensive experiments with various sampling strategies are conducted under different datasets. 
Based on our analysis, the continual learning problem certainly needs further research, since for complex datasets, even with advanced replay strategies, the accuracy remains low, and the forgetting rate remains high. 
This work would shed some light on selecting replay methods for various data distributions and lay a solid foundation for further theoretical studies.

\bibliographystyle{unsrt}
\bibliography{ijcai21.bib}

\clearpage
\onecolumn
\section{Appendix}
\begin{table}[H]
\caption{Replay Strategies Comparison Result Summary}
\label{table:Sum1}
\resizebox{\textwidth}{!}{
\begin{tabular}{@{}cccccccccc@{}}
\toprule
\multicolumn{2}{c}{} &
  \textbf{Random} &
  \begin{tabular}[c]{@{}c@{}}\textbf{Least}\\ \textbf{Entropy}\end{tabular} &
  \begin{tabular}[c]{@{}c@{}}\textbf{Largest}\\ \textbf{Confidence}\end{tabular} &
  \begin{tabular}[c]{@{}c@{}}\textbf{Largest}\\ \textbf{Margin}\end{tabular} &
  \begin{tabular}[c]{@{}c@{}}\textbf{Least Bayesian} \\ \textbf{Dis-agreement}\end{tabular} &
  \textbf{K-means} &
  \textbf{Core-Set} &
  \textbf{MIR} \\ \midrule \midrule
\textbf{} &
  Accuracy &
  $0.870$ +/- $0.012$ &
  $0.871$ +/- $0.014$ &
  $0.869$ +/- $0.009$ &
  $\textbf{0.876}$ +/- $\textbf{0.014}$ &
  $\textbf{0.876}$ +/- $\textbf{0.010}$ &
  $\textbf{0.876}$ +/- $\textbf{0.008}$ &
  $0.870$ +/- $0.010$ &
  $0.871$ +/- $0.014$ \\
\textbf{\begin{tabular}[c]{@{}c@{}}MNIST\end{tabular}} &
  \begin{tabular}[c]{@{}c@{}}Forgetting \\ rate\end{tabular} &
  $0.105$ +/- $0.015$ &
  $0.101$ +/- $0.019$ &
  $0.099$ +/- $0.014$ &
  $0.099$ +/- $0.017$ &
  $0.099$ +/- $0.014$ &
  $\textbf{0.096}$ +/- $\textbf{0.010}$ &
  $0.100$ +/- $0.013$ &
  $0.101$ +/- $0.018$ \\
 &
  \begin{tabular}[c]{@{}c@{}}Average time\\ (s)\end{tabular} &
  $3.050$ &
  $3.900$ &
  $\textbf{3.700}$ &
  $3.850$ &
  $5.150$ &
  $\textit{22.400}$ &
  $4.150$ &
  $4.350$ \\ 
  \midrule
\textbf{} &
  Accuracy &
  $0.339$ +/- $0.024$ &
  $0.393$ +/- $0.020$ &
  $0.402$ +/- $0.018$ &
  $0.400$ +/- $0.018$ &
  $0.395$ +/- $0.009$ &
  $0.349$ +/- $0.018$ &
  $0.345$ +/- $0.018$ &
  $\textbf{0.409}$ +/- $\textbf{0.013}$ \\
\textbf{\begin{tabular}[c]{@{}c@{}}CIFAR-10\end{tabular}} &
  \begin{tabular}[c]{@{}c@{}}Forgetting rate\\ \end{tabular} &
  $0.400$ +/- $0.040$ &
  $0.267$ +/- $0.036$ &
  $0.259$ +/- $0.036$ &
  $0.274$ +/- $0.035$ &
  $\textbf{0.221}$ +/- $\textbf{0.018}$ &
  $0.373$ +/- $0.031$ &
  $0.351$ +/- $0.026$ &
  $0.248$ +/- $0.019$ \\
 &
  \begin{tabular}[c]{@{}c@{}}Average time\\ (s)\end{tabular} &
  $62.400$ &
  $\textbf{89.200}$ &
  $90.067$ &
  $89.800$ &
  $234.333$ &
  $\textit{282.933}$ &
  $89.267$ &
  $152.133$ \\ 
  \midrule
\textbf{} &
  Accuracy &
  $0.107$ +/- $0.010$ &
  $0.116$ +/- $0.016$ &
  $0.134$ +/- $0.018$ &
  $0.134$ +/- $0.022$ &
  $\textbf{0.158}$ +/- $\textbf{0.014}$ &
  $0.111$ +/- $0.012$ &
  $0.109$ +/- $0.018$ &
  $0.126$ +/- $0.015$ \\
\textbf{\begin{tabular}[c]{@{}c@{}}MiniImagenet\end{tabular}} &
  \begin{tabular}[c]{@{}c@{}}Forgetting \\ rate\end{tabular} &
  $0.402$ +/- $0.012$ &
  $0.364$ +/- $0.013$ &
  $0.348$ +/- $0.026$ &
  $0.334$ +/- $0.027$ &
  $\textbf{0.280}$ +/- $\textbf{0.020}$ & 
  $0.382$ +/- $0.013$ &
  $0.358$ +/- $0.022$ &
  $0.338$ +/- $0.037$ \\
 &
  \begin{tabular}[c]{@{}c@{}}Average time\\ (s)\end{tabular} &
  $1183.000$ &
  $1290.800$ &
  $\textbf{1284.400}$ &
  $\textbf{1284.400}$ &
  $1622.800$ &
  $\textit{1745.800}$ &
  $1465.500$ &
  $1379.600$ \\
  \midrule
\textbf{} &
  Accuracy &
  $0.960$ +/- $0.000$ &
  $\textbf{0.966}$ +/- $\textbf{0.004}$ &
  $\textbf{0.966}$ +/- $\textbf{0.007}$ &
  $0.964$ +/- $0.011$ &
  $0.962$ +/- $0.004$ &
  $0.962$ +/- $0.004$ &
  $0.962$ +/- $0.004$ &
  $0.964$ +/- $0.007$ \\
\textbf{\begin{tabular}[c]{@{}c@{}}OpenLORIS\end{tabular}} &
  \begin{tabular}[c]{@{}c@{}}Forgetting \\ rate\end{tabular} &
  $0.022$ +/- $0.007$ &
  $\textbf{0.014}$ +/- $\textbf{0.004}$ &
  $0.020$ +/- $0.010$ &
  $0.018$ +/- $0.010$ &
  $0.024$ +/- $0.004$ &
  $0.016$ +/- $0.007$ &
  $0.026$ +/- $0.009$ &
  $0.020$ +/- $0.010$ \\
 &
  \begin{tabular}[c]{@{}c@{}}Average time\\ (s)\end{tabular} &
  $5521.800$ &
  $\textbf{5777.000}$ &
  $9582.000$ &
  $12017.400$ &
  $\textit{12270.400}$ &
  $7400.600$ &
  $8828.200$ &
  $6753.600$ \\ \bottomrule
\end{tabular}}
\end{table}

\begin{table}[H]
\caption{Replay Data Difficulty Comparison Result Summary}
\label{table:Sum2}
\resizebox{\textwidth}{!}{
\begin{tabular}{@{}cccccclcccc@{}}
\toprule
\multicolumn{2}{c}{\multirow{2}{*}{}} &
  \multicolumn{4}{c}{\textbf{Replay Simple Data}} &
   &
  \multicolumn{4}{c}{\textbf{Replay Difficult Data}} \\ \cmidrule(lr){3-6} \cmidrule(l){8-11} 
\multicolumn{2}{c}{} &
  \begin{tabular}[c]{@{}c@{}}Least\\ Entropy\end{tabular} &
  \begin{tabular}[c]{@{}c@{}}Largest\\ Confidence\end{tabular} &
  \begin{tabular}[c]{@{}c@{}}Largest\\ Margin\end{tabular} &
  \begin{tabular}[c]{@{}c@{}}Least Bayesian \\ Disagreement\end{tabular} &
   &
  \begin{tabular}[c]{@{}c@{}}Largest\\ Entropy\end{tabular} &
  \begin{tabular}[c]{@{}c@{}}Least\\ Confidence\end{tabular} &
  \begin{tabular}[c]{@{}c@{}}Least\\ Margin\end{tabular} &
  \begin{tabular}[c]{@{}c@{}}Largest Bayesian\\ Disagreement\end{tabular} \\ \midrule \midrule
\textbf{} &
  Accuracy &
  $0.871$ +/- $0.014$ &
  $0.869$ +/- $0.009$ &
  $\textbf{0.876}$ +/- $\textbf{0.014}$ &
  $\textbf{0.876}$ +/- $\textbf{0.010}$ &
   &
  $\textbf{0.877}$ +/- $\textbf{0.008}$ &
  $\textbf{0.874}$ +/- $\textbf{0.009}$ &
  $0.872$ +/- $0.014$ &
  $0.876$ +/- $0.009$ \\
\textbf{\begin{tabular}[c]{@{}c@{}}MNIST\end{tabular}} &
  \begin{tabular}[c]{@{}c@{}}Forgetting \\ rate\end{tabular} &
  $0.101$ +/- $0.019$ &
  $0.099$ +/- $0.014$ &
  $\textbf{0.099}$ +/- $\textbf{0.017}$ &
  $\textbf{0.099}$ +/- $\textbf{0.014}$ &
   &
  $\textbf{0.095}$ +/- $\textbf{0.010}$ &
  $\textbf{0.098}$ +/- $\textbf{0.016}$ &
  $0.100$ +/- $0.019$ &
  $0.101$ +/- $0.014$ \\
 &
  \begin{tabular}[c]{@{}c@{}}Average time\\ (s)\end{tabular} &
  $3.900$ &
  $\textbf{3.700}$ &
  $\textbf{3.850}$ &
  $5.150$ &
   &
  $\textbf{3.700}$ &
  $4.000$ &
  $4.100$ &
  $\textbf{4.750}$ \\ 
  \midrule
\textbf{} &
  Accuracy &
  $0.393$ +/- $0.020$ &
  $0.402$ +/- $0.018$ &
  $0.400$ +/- $0.018$ &
  $0.395$ +/- $0.009$ &
   &
  $\textbf{0.402}$ +/- $\textbf{0.017}$ &
  $\textbf{0.403}$ +/- $\textbf{0.011}$ &
  $\textbf{0.407}$ +/- $\textbf{0.013}$ &
  $\textbf{0.403}$ +/- $\textbf{0.008}$ \\
\textbf{\begin{tabular}[c]{@{}c@{}}CIFAR-10\end{tabular}} &
  \begin{tabular}[c]{@{}c@{}}Forgetting \\ rate\end{tabular} &
  $0.267$ +/- $0.036$ &
  $0.259$ +/- $0.036$ &
  $0.274$ +/- $0.035$ &
  $0.221$ +/- $0.018$ &
   &
  $\textbf{0.244}$ +/- $\textbf{0.028}$ &
  $\textbf{0.233}$ +/- $\textbf{0.023}$ &
  $\textbf{0.247}$ +/- $\textbf{0.017}$ &
  $\textbf{0.211}$ +/- $\textbf{0.018}$ \\
 &
  \begin{tabular}[c]{@{}c@{}}Average time\\ (s)\end{tabular} &
  $\textbf{89.200}$ &
  $\textbf{90.067}$ &
  $89.800$ &
  $\textbf{234.333}$ &
   &
  $89.933$ &
  $91.667$ &
  $\textbf{88.733}$ &
  $235.800$ \\
  \midrule
\textbf{} &
  Accuracy &
  $0.116$ +/- $0.016$ &
  $0.134$ +/- $0.018$ &
  $0.134$ +/- $0.022$ &
  $\textbf{0.158}$ +/- $\textbf{0.014}$ &
   &
  $\textbf{0.122}$ +/- $\textbf{0.007}$ &
  $\textbf{0.146}$ +/- $\textbf{0.018}$ &
  $\textbf{0.136}$ +/- $\textbf{0.029}$ &
  $0.148$ +/- $0.017$ \\
\textbf{\begin{tabular}[c]{@{}c@{}}MiniImagenet\end{tabular}} &
  \begin{tabular}[c]{@{}c@{}}Forgetting \\ rate\end{tabular} &
  $0.364$ +/- $0.013$ &
  $0.348$ +/- $0.026$ &
  $\textbf{0.334}$ +/- $\textbf{0.027}$ &
  $0.280$ +/- $0.020$ & 
   &
  $\textbf{0.358}$ +/- $\textbf{0.004}$ &
  $\textbf{0.322}$ +/- $\textbf{0.031}$ &
  $0.336$ +/- $0.025$ &
  $\textbf{0.260}$ +/- $\textbf{0.032}$ \\
 &
  \begin{tabular}[c]{@{}c@{}}Average time\\ (s)\end{tabular} &
  $1290.800$ &
  $1284.400$ &
  $1284.400$ &
  $1622.800$ &
   &
  $\textbf{1266.000}$ &
  $\textbf{1267.000}$ &
  $\textbf{1280.400}$ &
  $\textbf{1607.200}$ \\ \bottomrule
\end{tabular}}
\end{table}

\end{document}